\documentclass[10pt,twocolumn,letterpaper]{article}

\usepackage{cvpr}
\usepackage{times}
\usepackage{epsfig}
\usepackage{graphicx}
\usepackage{amsmath}
\usepackage{amssymb}
\usepackage{multirow}
\usepackage{mathrsfs}

\usepackage[breaklinks=true,bookmarks=false]{hyperref}

\cvprfinalcopy 

\def\cvprPaperID{****} 

\begin{document}

\title{FontGAN: A Unified Generative Framework for Chinese Character Stylization and De-stylization}

\author{Xiyan Liu$^{1,2}$, Gaofeng Meng$^1$, Shiming Xiang$^1$ and Chunhong Pan$^1$ \\
$^1$National Laboratory of Pattern Recognition, Institute of Automation, Chinese Academy of Sciences,\\
Beijing 100190, China \\
$^2$School of Artificial Intelligence, University of Chinese Academy of Sciences,\\
Beijing 100049, China \\
{\tt\small \{xiyan.liu, gfmeng, smxiang, chpan\}@nlpr.ia.ac.cn}
}

\maketitle

\begin{abstract}
   Chinese character synthesis involves two related aspects, i.e., style maintenance and content consistency. Although some methods have achieved remarkable success in synthesizing a character with specified style from standard font, how to map characters to a specified style domain without losing their identifiability remains very challenging. In this paper, we propose a novel model named FontGAN, which integrates the character stylization and de-stylization into a unified framework. In our model, we decouple character images into style representation and content representation, which facilitates more precise control of these two types of variables, thereby improving the quality of the generated results. We also introduce two modules, namely, font consistency module (FCM) and content prior module (CPM). FCM exploits a category guided Kullback-Leibler loss to embedding the style representation into different Gaussian distributions. It constrains the characters of the same font in the training set globally. On the other hand, it enables our model to obtain style variables through sampling in testing phase. CPM provides content prior for the model to guide the content encoding process and alleviates the problem of stroke deficiency during de-stylization. Extensive experimental results on character stylization and de-stylization have demonstrated the effectiveness of our method.
\end{abstract}


\section{Introduction}

Unlike English characters, which consist of only 26 alphabets, the number of Chinese characters is quite large. According to the official statistics, the total number of Chinese characters is 91251 and the frequently-used Chinese characters is 3500. For such a large number of characters, manually designing a new font is always time-consuming and error prone. Therefore, it is very important and meaningful to design a model to automatically generate Chinese characters with specified font style and explicit semantic information.

Previous studies have shown significant progresses on character generation. A large amount of approaches~\cite{xu2009automatic,lian2012automatic,lian2016automatic,zong2014strokebank} focus on analyzing and extracting stroke features, while some methods~\cite{zi2zi,Rewrite,zhang2018separating,chang2018generating} treat font transfer as an image-to-image task. However, there are still some challenges, which are in three folds. Firstly, unlike typical image generation tasks, Chinese character synthesis is very sensitive to changes in content. Any slight change may lead to a change in the meaning of the character. Secondly, large topological variations make the translation between different fonts more challenging. Finally, due to the large number of font types, it is impractical to learn a mapping function between each of the two fonts. Therefore, how to implement multiple font translations with one framework is particularly important.

The aforementioned challenges motivate us to propose an effective method to automatically generate more accurate and stable characters. We formulate the process of transferring standard font (e.g. SimSun) to other fonts as character stylization. On the contrary, the character de-stylization is defined as mapping various fonts to standard font (e.g. SimSun). Unlike character recognition, character de-stylization provides a new idea to normalize various fonts to a standard font, which is based on image-to-image translation. We think it will facilitate text digitization and subsequent editing and sharing.

Specifically, as illustrated in Fig.~\ref{fig:2}, our model takes two images of different fonts and contents as input and decouples them into style representation and content representation, respectively. We then exchange these two sets of variables, and finally generate characters that match the specified style and content. In addition, we incorporate character stylization and de-stylization in a universal framework, which allows two tasks to be trained simultaneously. Moreover, we introduce font consistency module (FCM) to encode various styles to their respective Gaussian distributions. On one hand, it constrains the training dataset globally, which promotes the style variable of the same font belong to the same distribution, and different fonts conform to different distributions. On the other hand, in the testing phase, it not only allows to obtain the style variable by encoding the reference image using style encoder, but also via sampling. In addition, compared to standard font, the topology of some calligraphy or handwritten fonts changes significantly. Therefore, when performing character de-stylization, the generated characters often lack strokes or strokes are confusing. In order to alleviate this problem, we pre-trained a model called content prior module (CPM) to provide additional constraints for the content encoder. CPM can encourage the encoder to be optimized in the direction that is easier to synthesize standard font.

The main contributions can be summarized as follows.

\begin{itemize}
\item We propose a novel unified framework called FontGAN for modeling Chinese character stylization and de-stylization together. Different from character recognition, character de-stylization provides a new way to normalize multiple fonts into one standard font, which will facilitate text digitization and subsequent editing and sharing.
\item We decouple the image into the style representation and content representation. A content prior module (CPM) is designed to improve the stability and accuracy of the character de-stylization. In addition, we propose font consistency module (FCM) to encode the same font into the same Gaussian distribution, which allows us to obtain the specified font style by sampling during testing.
\item Our model can handle one-to-many, many-to-one and many-to-many character generation tasks. Extensive experiments are performed to demonstrate the superior performance of our method over the state-of-the-art method.
\end{itemize}

\section{Related work}

\subsection{Generative Adversarial Network}

Generative adversarial network (GAN)~\cite{GAN} generally consists of two parts: generator and discriminator. With adversarial training, the discriminator enforces the generator to capture the distribution of real data. In recent years, GAN has achieved impressive results in various applications such as image synthesis, image super-resolution, domain adaptation, etc. Based on GAN, conditional GAN (CGAN)~\cite{mirza2014conditional} was proposed to provide additional guidance information, which encourages GAN to generate samples in our desired direction. There are many excellent CGAN-based structures that conditioned on discrete labels~\cite{odena2017conditional,li2019global}, images~\cite{radford2015unsupervised,cyclegan,zhu2017toward}, texts~\cite{reed2016generative,zhang2017stackgan,liu2018semantic} and so on. Pix2pix~\cite{pix2pix} is a classical CGAN-based framework for image-to-image translation, which applies paired data, skip-connection and patchGAN to achieve high-quality results. Pix2pix has demonstrated competitive performance in a variety of image translation tasks including sketch-to-photo, semantic segmentation, colorization, etc. In our work, we design a CGAN-based structure and treat character stylization and de-stylization as image-to-image translation task.

\subsection{Disentangled Representation Learning}

One of the goals of disentangled representation learning is to disentangle the underlying factors of data variation. There exists a great deal of literature in this field. InfoGAN~\cite{infogan} learns the meaningful representations by maximizing the mutual information between the latent codes and generated images. Tran et al.~\cite{tran} propose DR-GAN and explicitly disentangle the identity representation for pose-invariant face recognition and face synthesis. DRIT~\cite{lee} achieve unpaired diverse image-to-image translation by decomposing the input image into domain-invariant content space and domain-specific attribute space. Benefiting from the disentangled representation learning, character can be disentangled into content-related code and font-related distribution, which helps to learn different components better.

\subsection{Chinese Character Synthesis}

Many works on Chinese character synthesis~\cite{xu2009automatic,lian2012automatic,lian2016automatic,zong2014strokebank,miyazaki2017automatic} rely on stroke extraction. Xu et al.~\cite{xu2009automatic} synthesize handwriting style images by analyzing the strike shapes and character topology. Zong et al.~\cite{zong2014strokebank} present StrokeBank, which is a dictionary that maps components in standard font to a particular handwriting. Lian et al.~\cite{lian2016automatic} aim to provide an effective system to generate personal handwriting font library. In their work, stroke attributes are learned by Artificial Neural Networks (ANNs). Furthermore, this system only needs a small number of samples that are carefully written by a person. In addition, some methods~\cite{zi2zi,Rewrite,lyu2017auto,zhang2018separating,chang2018generating} treat font synthesis as an image-to-image translation task. Zi2zi~\cite{zi2zi}is an excellent GAN-based model for one-to-many font generation and achieves the state-of-the-art performance. It applies pix2pix structure and category embedding to generate the desired font conditioned on discrete label. A classification module is added to discriminator to calculate the category loss.

\section{Approach}

\begin{figure*}
\begin{center}
\includegraphics[width=0.9\linewidth]{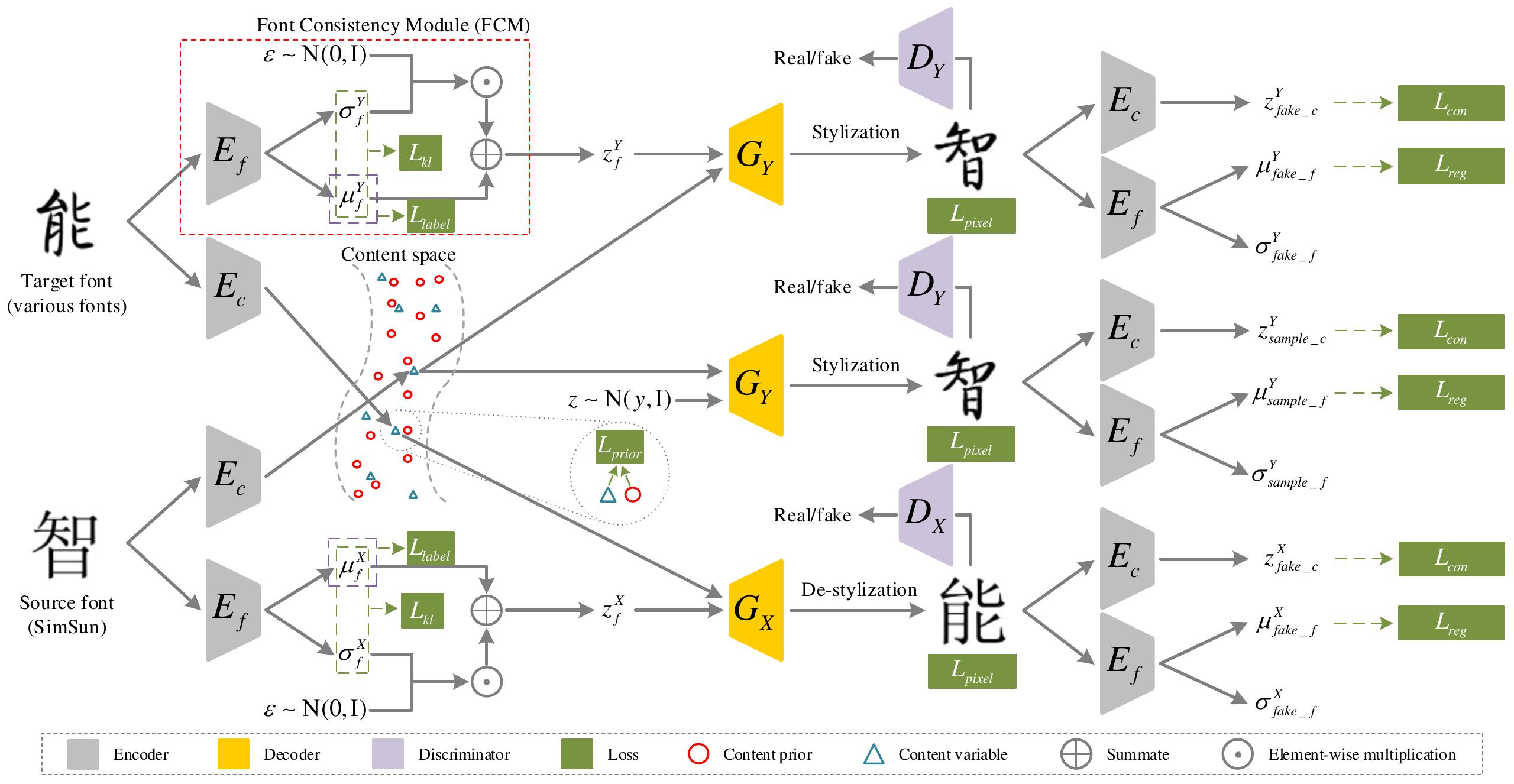}
\end{center}
   \caption{The architecture of FontGAN, which consists of content encoder ($E_{c}$), style encoder ($E_{f}$), decoder ($G_{X}, G_{Y}$) and discriminator ($D_{X}, D_{Y}$). $E_{c}$ and $E_{f}$ map the character into content variable and style variable, respectively. Then, exchange each other's variables and feed them into decoder to generate new character, which correspond to the specified content and style. The discriminators are used to distinguish whether the input is a real character or a generated image. In addition, the content prior is used to provide additional constraints on content space. Moreover, the KL-loss enforces the style variable to their respective Gaussian distributions, which allows us to obtain style information by sampling during testing.}
\label{fig:2}
\end{figure*}

\subsection{Model Overview}

\textbf{Notation:} Let $I^{X}$, $I^{Y}$ denote the source font image and target font image, respectively. Moreover, we use $I_{ref}^{X}$ and $I_{ref}^{Y}$ denote the reference image for pixel-wise loss.
We define the proposed subnets as follows: content encoder ($E_{c}$), style encoder ($E_{f}$), stylization decoder ($G_{Y}$), de-stylization decoder ($G_{X}$), target font discriminator ($D_{Y}$) and source font discriminator ($D_{X}$).

The proposed FontGAN follows a typical encoder-decoder architecture for character synthesis. In the framework, we introduce two different types of encoder, i.e., content encoder and style encoder, as shown in Fig.~\ref{fig:2}. The source font and target font share the content encoder and style encoder. This is because the content is style-irrelevant, and the style encoder has the ability to learn the representation of various fonts.

Specifically, during the training phase, the input images are first encoded into latent variables as follows:
\begin{equation}
\begin{split}
z_{c}^{X} = E_{c}(I^{X}), \quad z_{f}^{X} = E_{f}(I^{X})
\end{split}
\label{eq:1}
\end{equation}
\begin{equation}
\begin{split}
z_{c}^{Y} = E_{c}(I^{Y}), \quad z_{f}^{Y} = E_{f}(I^{Y})
\end{split}
\label{eq:2}
\end{equation}
where $z_{c}^{X}$ and $z_{f}^{X}$ denote the content variable and style variable in $X$ domain, $z_{c}^{Y}$ and $z_{f}^{Y}$ denote the content variable and style variable in $Y$ domain, respectively.

The content variables and style variables are then exchanged and combined. The combined variables are finally fed into decoder to generate images as below:
\begin{equation}
\begin{split}
I_{fake}^{X} = G_{X}(z_{c}^{Y}, z_{f}^{X})
\end{split}
\label{eq:3}
\end{equation}
\begin{equation}
\begin{split}
I_{fake}^{Y} = G_{Y}(z_{c}^{X}, z_{f}^{Y})
\end{split}
\label{eq:4}
\end{equation}
where $I_{fake}^{X}$ represents the generated image, which has the content of $I^{Y}$ and the font of $I^{X}$. Similarly, $I_{fake}^{Y}$ is the generated image, which has the content of $I^{X}$ and the font of $I^{Y}$.

After generation, the synthetic images are distinguished by discriminator. This mechanism mainly has two advantages: 1) It incorporates character stylization and de-stylization in one framework; 2) Content representation and style representation are explicitly disentangled.

During the testing phase, we have two ways to generate the desired image. One is that the style encoder encodes the input image into style variable, and then concatenated with the content variable to get the final result. The other is that we can sample from Gaussian distribution to obtain the style variable, which will be described in detail in the following subsection.

\subsection{Network Architecture}

Our model mainly consists of two parts: the generator and the discriminator (as shown in Fig.~\ref{fig:2}). Furthermore, generator contains four subnets: content encoder, style encoder, stylization decoder and de-stylization decoder. In addition, we also design two discriminators for target font and source font. Next, we will introduce these networks in detail.

\subsubsection{Encoder Network}

In order to separate the content variable and style variable, we propose two networks: content encoder and style encoder. These networks have similar structure except the last module. Specifically, the encoder has 5 down-sampling modules to down sample the input image with stride-2. Each down-sampling module is composed of a convolutional layer and a residual block~\cite{37}, wherein the convolutional layer is used to implement the downsampling operation. Different from content encoder, style encoder has a fully connected layer after the final down-sampling module. This fully connected layer is used to obtain $\mu$ (128-dim) and $\sigma$ (128-dim) for Gaussian sampling.

\subsubsection{Decoder Network}

Stylization decoder and de-stylization decoder map the combined variables to the final images. These decoders are structurally consistent. Specifically, the decoder network is composed of 4 up-sampling modules, which consists of a deconvolutional layer with stride-2 and a residual block. Moreover, skip-connection~\cite{ronneberger2015u} is applied to refine the generated results.

\subsubsection{Discriminator Network}

In our approach, we feed the discriminator $D_{Y}$ with two types of input pairs: $cat(I^{X}, I_{ref}^{Y})$ for positive example, $cat(I^{X}, I_{fake}^{Y})$ for negative example. The discriminator $D_{X}$ is similar to $D_{Y}$. In practice, the discriminator takes the advantages of the previous method~\cite{pix2pix} and consists of a series of down-sampling modules. It down samples the input image to $4 \times 4$ spatially, and then tries to classify if each $4 \times 4$ patch is real or fake.

\subsection{Loss functions}

\subsubsection{Adversarial Loss}

In order to generate realistic and clear images, we apply adversarial loss to train our model. The generator tries to generate the desired images by minimizing the following objective, while the discriminator aims to distinguish the real image and the synthetic image by maximizing the objective. The objective of the adversarial framework can be written as:
\begin{equation}
\begin{split}
&\mathcal{L}_{GAN}(E_{c}, E_{f}, G_{Y}, D_{Y})\\
&=\mathbb{E}_{(I^{X}, I_{ref}^{Y})\sim P_{data}}\lbrack\log D_{Y}(<I^{X}, I_{ref}^{Y}>)\rbrack\\
&+\mathbb{E}_{(I^{X}, I_{fake}^{Y})\sim P_{data}}\lbrack\log(1-D_{Y}(<I^{X}, I_{fake}^{Y}>))\rbrack
\end{split}
\label{eq:5}
\end{equation}

\begin{equation}
\begin{split}
&\mathcal{L}_{GAN}(E_{c}, E_{f}, G_{X}, D_{X})\\
&=\mathbb{E}_{(I^{Y}, I_{ref}^{X})\sim P_{data}}\lbrack\log D_{X}(<I^{Y}, I_{ref}^{X}>)\rbrack\\
&+\mathbb{E}_{(I^{Y}, I_{fake}^{X})\sim P_{data}}\lbrack\log(1-D_{X}(<I^{Y}, I_{fake}^{X}>))\rbrack
\end{split}
\label{eq:6}
\end{equation}

\begin{equation}
\begin{split}
&\mathcal{L}_{GAN}(E_{c}, G_{Y}, D_{Y})\\
&=\mathbb{E}_{(I^{X}, I_{ref}^{Y})\sim P_{data}}\lbrack\log D_{Y}(<I^{X}, I_{ref}^{Y}>)\rbrack\\
&+\mathbb{E}_{(I^{X}, I_{sam}^{Y})\sim P_{data}}\lbrack\log(1-D_{Y}(<I^{X}, I_{sam}^{Y}>))\rbrack
\end{split}
\label{eq:7}
\end{equation}
where $<\cdot>$ means concatenating at channel level. $I_{sam}^{Y}=G^{Y}(z_{c}^{X}, z_{sample})$, where $z_{sample}$ is obtained by sampling the Gaussian distribution corresponding to the target font.

\subsubsection{Pixel-wise Loss}

Different from typical image generation tasks, Chinese character synthesis has higher requirements for the integrity and accuracy of the generated content. Any change will cause the character to have no practical meaning or change its own semantics. Therefore, we suggest to impose pixel-wise loss to constrain the generated results.

\begin{equation}
\begin{split}
\mathcal{L}_{pixel} &= \|I_{fake}^{Y}-I_{ref}^{Y}\|_{2}^{2} + \|I_{sam}^{Y}-I_{ref}^{Y}\|_{2}^{2} \\
&+ \|I_{fake}^{X}-I_{ref}^{X}\|_{2}^{2}
\end{split}
\label{eq:8}
\end{equation}

We use L2 loss for image synthesis in our approach. Because L2 loss is more sensitive to pixel-level changes than L1 loss, which makes it easier to guarantee the integrity of the generated characters.

\subsubsection{Content Consistent Loss}

During the training phase, the content encoder will map the input image into a shared content space. And the content of the generated image should be consistent with the original image. Therefore, we formulate the content consistent loss as:
\begin{equation}
\begin{split}
\mathcal{L}_{con} &= \|z_{fake\_c}^{X}-z_{c}^{Y}\|_{2}^{2} + \|z_{fake\_c}^{Y}-z_{c}^{X}\|_{2}^{2} \\
&+\|z_{sam\_c}^{Y}-z_{c}^{X}\|_{2}^{2}
\end{split}
\label{eq:9}
\end{equation}
where $z_{fake\_c}^{X}$, $z_{fake\_c}^{Y}$ and $z_{sam\_c}^{Y}$ are the outputs of the $E_{c}$ with $I_{fake}^{X}$, $I_{fake}^{Y}$ and $I_{sam}^{Y}$ as inputs.

\subsubsection{Category-guided KL Loss}

We introduce font consistency module (FCM) to decompose the style variables from character. Inspired by the original VAE~\cite{kingma2013auto}, the optimization objective for our style encoder $E_{f}$ is to maximize the lower bound of $\log p_{\theta}(x)$, which can be written as:
\begin{equation}
\begin{split}
\log p_{\theta}(x) \geq &-D_{KL}(q_{\phi}(z_{f}|x)||p_{\theta}(z_{f}))\\
& + E_{q_{\phi}(z_{f}|x)}[\log p_{\theta}(x|z_{f})]
\end{split}
\label{eq:10}
\end{equation}
where $\phi$ is the variational parameters, and $\theta$ represents the generative parameters. Moreover, $z_{f}$ denotes the style variables, and $p_{\theta}(z_{f})$ is the prior distributions for $z_{f}$.

In practice, we assume the prior over the style variable is the centered isotropic multivariate Gaussian, which can be formulate as $p_{\theta}(z_{f})=\mathcal{N}(\textbf{y}, \textbf{I})$, where $\textbf{y}$ represents the vector filled by the font label $y$. In our experiments, $y=0$ represents the source font SimSun, and $y \in (1, 2, \cdots, n)$ for other target fonts, where $n$ is the number of fonts. In this way, on one hand, it ensures that the style variables of the same fonts can be encoded into the same Gaussian distribution. On the other hand, in the testing phase, in addition to directly encoding the reference image to get style variables, it can also be obtained by sampling.

Specifically, we can optimize Eq.(\ref{eq:10}) by minimizing the following KL-loss:
\begin{equation}
\begin{split}
\mathcal{L}_{s\_kl}= \frac{1}{2}\sum_{i}^{d}((\mu_{f}^{i})^{2}+(\sigma_{f}^{i})^{2}-\log((\sigma_{f}^{i})^{2})-1)
\end{split}
\label{eq:11}
\end{equation}
\begin{equation}
\begin{split}
\mathcal{L}_{t\_kl}= \frac{1}{2}\sum_{i}^{d}((\mu_{f}^{i}-y)^{2}+(\sigma_{f}^{i})^{2}-\log((\sigma_{f}^{i})^{2})-1)
\end{split}
\label{eq:12}
\end{equation}
where $\mathcal{L}_{s\_kl}$ and $\mathcal{L}_{t\_kl}$ denote the source font KL-loss and the target font KL-loss, respectively. In addition, $\mu_{f}$ and $\sigma_{f}$ are the outputs of the style encoder $E_{f}$. The style variable $z_{f}$ is sampled from $\mathcal{N}(z_{f}; \mu_{f}, \sigma_{f}^{2})$ using $z_{f}=\mu_{f}+\sigma_{f} \odot \epsilon_{f}$, where $\epsilon_{f} \sim \mathcal{N}(\textbf{0}, \textbf{I})$. Moreover, $d$ is the dimension of $z_{f}$.

\begin{figure}
\begin{center}
\includegraphics[width=1.0\linewidth]{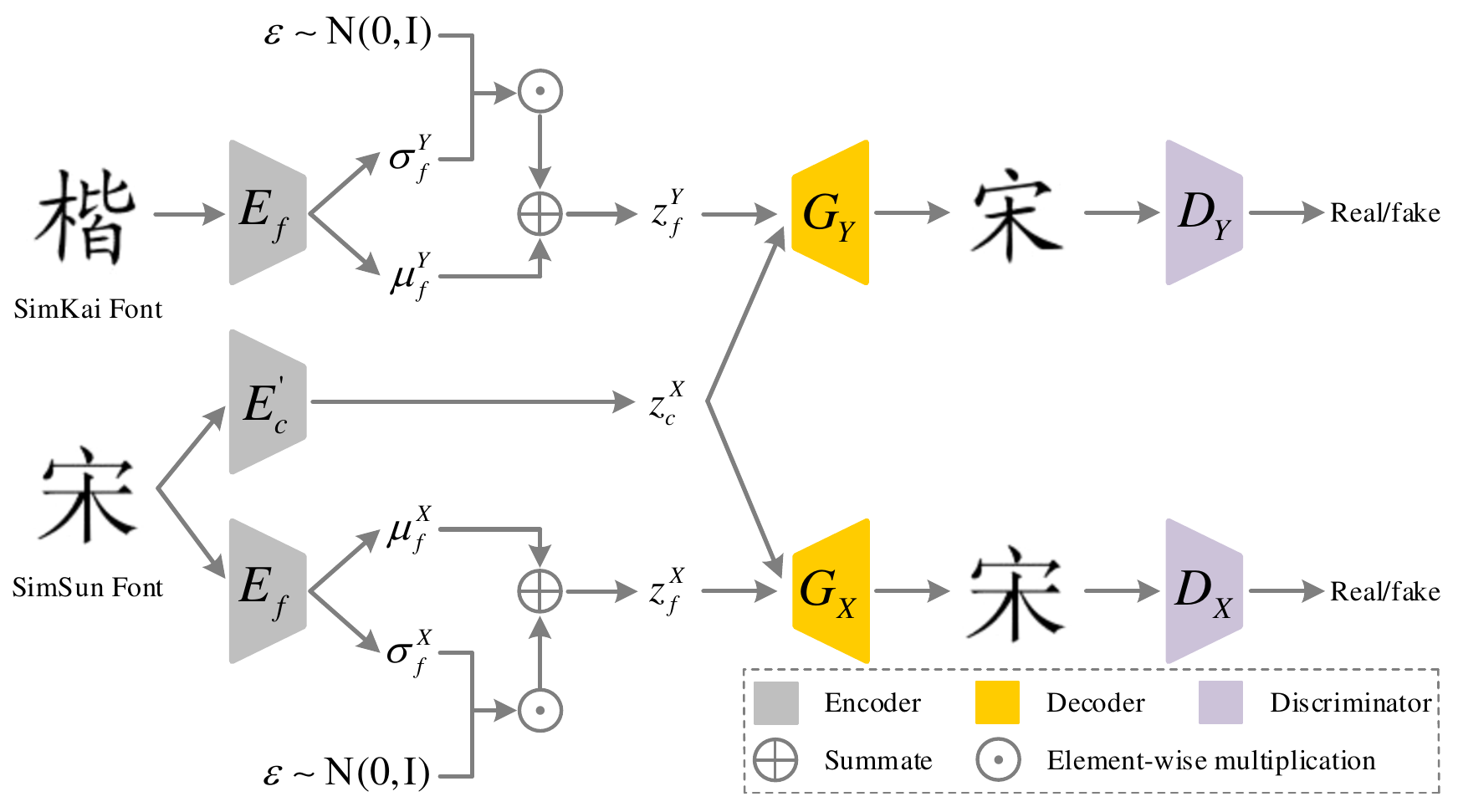}
\end{center}
   \caption{The structure of Content Prior Module (CPM). By introducing the SimKai font, the content variable of SimSun can be accurately decomposed. We use $E_{c}^{'}$ to provide content prior to guide the training of $E_{c}$ in the final model.}
\label{fig:4}
\end{figure}

\subsubsection{Font Label Preserving Loss}

In order to further disentangle the style variable of different fonts, we apply the label preserving loss to constrain the style encoder $E_{f}$ to encode different fonts to their respective distributions more accurately. We compute this loss function as follows:
\begin{equation}
\begin{split}
\mathcal{L}_{label}= |\mu_{f}^{X}|+|\mu_{f}^{Y}-y|
\end{split}
\label{eq:13}
\end{equation}
where $\mu_{f}$ is the output of $E_{f}$, $y \in (1, 2, \cdots, n)$ is the font label.

\subsubsection{Latent Regression Loss}

Inspired by~\cite{lee,zhu2017toward}, we believe that the generated image should be consistent with the font of the image that provides style information for it. Therefore, we construct a L1 loss to regress the style latent variable. It can be defined as:
\begin{equation}
\begin{split}
\mathcal{L}_{reg}= |\hat{z}_{f}^{X}-z_{f}^{X}| + |\hat{z}_{f}^{Y}-z_{f}^{Y}| + |\hat{z}_{sam\_f}^{Y}-z_{f}^{Y}|
\end{split}
\label{eq:14}
\end{equation}
where $z_{f}^{X}=E_{f}(I^{X})$ and $z_{f}^{Y}=E_{f}(I^{Y})$. Furthermore, $\hat{z}_{f}^{X}=E_{f}(I_{fake}^{X})$, $\hat{z}_{f}^{Y}=E_{f}(I_{fake}^{Y})$, $\hat{z}_{sam\_f}^{Y}=E_{f}(I_{sam}^{Y})$.

\subsubsection{Content Prior Loss}

With regard to the de-stylization process, the generated results are often unsatisfactory due to the large difference in topology between the target font and the standard font. Therefore, we propose a content prior module (CPM) (as shown in Fig.~\ref{fig:4}), which is used to provide content prior for our framework. Specifically, firstly, we pre-train a simple model that is similar to our final structure. SimSun and SimKai are treated as source font and target font, since they are the most common fonts, and the layout is very simple. Using the loss function mentioned above, we can get a trained content encoder $E_{c}^{'}$. Secondly, when we train our final model, $E_{c}^{'}$ will guides $E_{c}$ to encode the content of the target font into the content prior space, where the content variables in the space are more likely to generate satisfactory SimSun characters using $I_{fake}^{X}=G^{X}(E_{c}(I^{Y}), z_{f}^{X})$. Finally, the loss function can be written as:
\begin{equation}
\begin{split}
\mathcal{L}_{prior}=\|E_{c}(I^{Y})-E_{c}^{'}(I_{ref}^{X})\|_{2}^{2}
\end{split}
\label{eq:15}
\end{equation}

To sum up, the full objective of our model can be expressed as:
\begin{equation}
\begin{split}
&~\mathcal{L}(E_{c}, E_{f}, G_{X}, D_{X}, G_{Y}, D_{Y})\\
& =\mathcal{L}_{GAN}(E_{c}, E_{f}, G_{Y}, D_{Y}) + \mathcal{L}_{GAN}(E_{c}, E_{f}, G_{X}, D_{X}) \\
& + \mathcal{L}_{GAN}(E_{c}, G_{Y}, D_{Y}) + \lambda_{pix}\mathcal{L}_{pixel} + \lambda_{con}\mathcal{L}_{con} \\
& + \lambda_{kl}(\mathcal{L}_{s\_kl} + \mathcal{L}_{t\_kl}) + \lambda_{lab}\mathcal{L}_{label} + \lambda_{reg}\mathcal{L}_{reg} + \lambda_{pri}\mathcal{L}_{prior}
\end{split}
\label{eq:16}
\end{equation}
where $\lambda$ controls the relative importance of the above objectives.

\section{Experiments}

In this section, we first introduce the implementation details, datasets, baseline methods and evaluation metrics. Then, extensive qualitative and quantitative experiments are conducted to validate the effectiveness of our approach. In addition, we conduct a series of experiments including the transfer between arbitrary fonts and the inference with novel fonts. Finally, we perform the ablation experiments to demonstrate the essentials of each components of our model.

\subsection{Experiment Setup}

\subsubsection{Implementation Details}

We have trained the network 100 epochs using Adam Optimizer with batch size 32 on TITAN XP GPU. The learning rate is initialized to 0.0002 and is reduced by half every 20 epochs. In our experiments, the parameters in Eq.(\ref{eq:16}) are fixed at $\lambda_{pix}=30$, $\lambda_{con}=\lambda_{pri}=5$, $\lambda_{lab}=\lambda_{reg}=1$, $\lambda_{kl}=0.01$. Moreover, all images are resized to $64 \times 64$, and are normalized into $[-1,1]$.

\subsubsection{Datasets}

As for content prior module, we collect SimSun and SimKai to train this module, and each font consisting of 6000 characters. For FontGAN, because there are no existing public datasets for Chinese character synthesis, we build a dataset that includes 50 fonts and each font contains 2000 Chineses characters. In addition, all characters in the test set have never appeared in the training set.

\subsubsection{Baseline Methods}

Four previous methods are adopted as baselines, including Rewrite, pix2pix, zi2zi and CGAN. Rewrite combines convolutional layer and maxpooling layer to implement translation between two fonts. Pix2pix is a commonly used model in the field of image translation. Zi2zi is the state-of-the-art method in Chinese font translation, which applies pix2pix structure and category embedding to generate the target font. Conditional GAN (CGAN) takes source font image as input and uses the one-hot vector as condition to generate character. In our experiments, CGAN uses FontGAN network as the main framework.

\subsubsection{Evaluation Metrics}

We use Multi-Scale Structural Similarity (MS-SSIM)~\cite{msssim}, Local Distortion (LD)~\cite{ld}, L1 loss and OCR accuracy as the evaluation metrics. MS-SSIM is a widely used image evaluation metric to measure the similarity between two images. LD is adopted to evaluate local distortion via dense SIFT flow. L1 loss can calculate pixel-level deviations. OCR is used to quantify the de-stylization results. For MS-SSIM and OCR accuracy, higher value means better performance, while LD and L1 loss are opposite.

\subsection{Qualitative Evaluation}

\subsubsection{Stylization in Simple Cases}

\begin{figure}
\begin{center}
\includegraphics[width=1.0\linewidth]{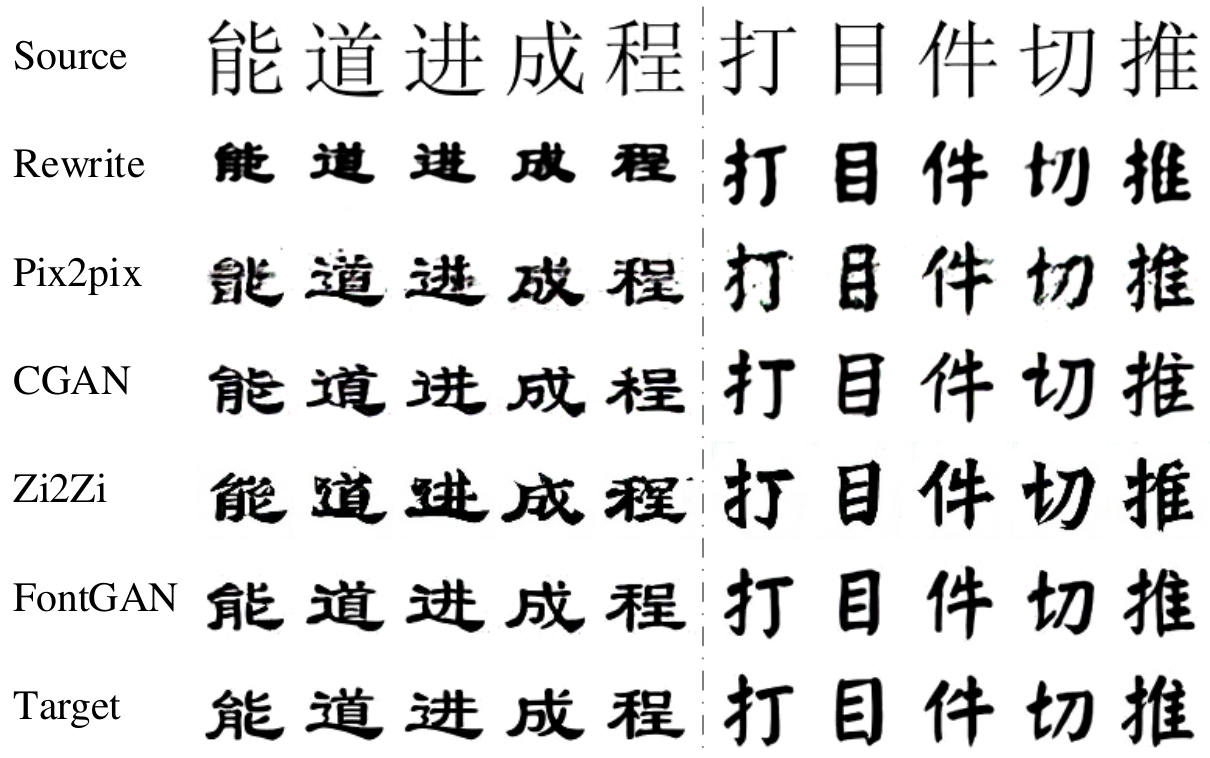}
\end{center}
   \caption{Character stylization comparisons in simple cases. Two font styles are displayed, each consisting of five characters.}
\label{fig:5}
\end{figure}

Fig.~\ref{fig:5} illustrates the stylization results in simple cases (e.g. the printed fonts). Since these fonts are standard and there are few joined-up characters, most of the methods can successfully transfer SimSun font into target fonts. It is worth noting that our method learns the style of the font more accurately.

\subsubsection{Stylization with Large Topology Change}

\begin{figure}
\begin{center}
\includegraphics[width=1.0\linewidth]{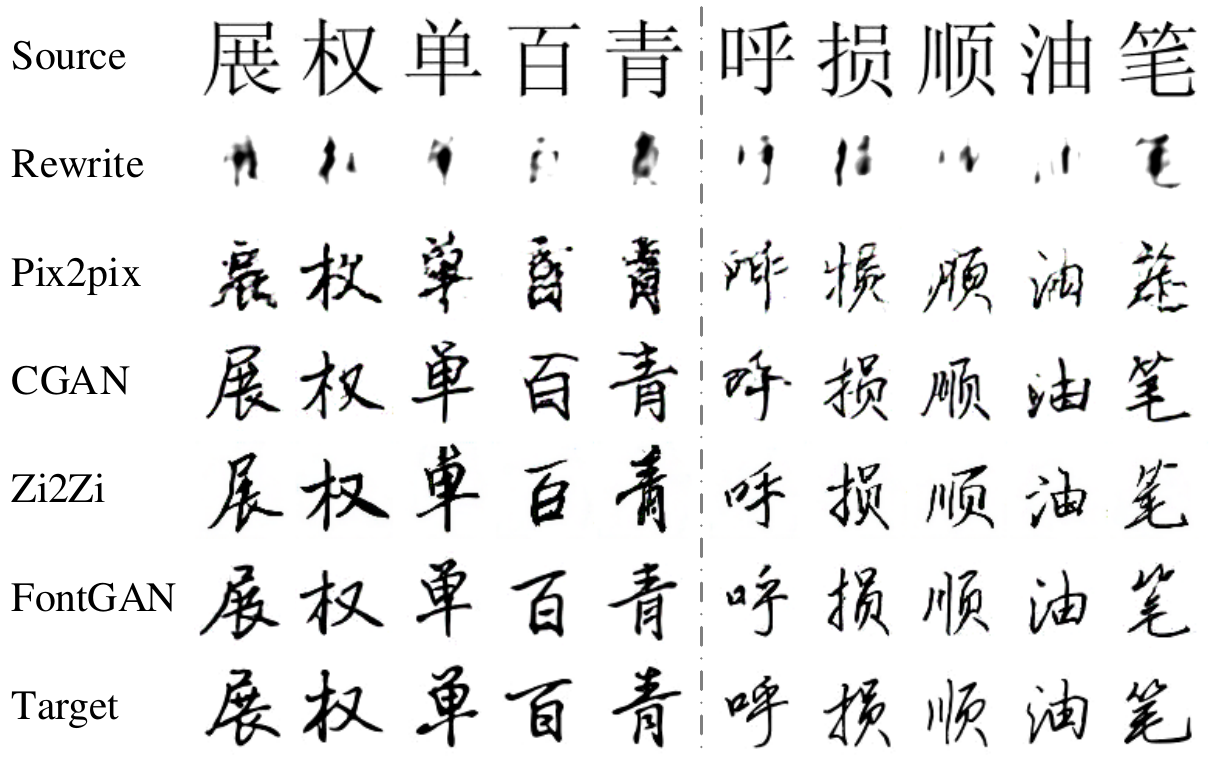}
\end{center}
   \caption{Character stylization comparisons in challenging cases. Two font styles are displayed, each consisting of five characters.}
\label{fig:6}
\end{figure}

\begin{figure}
\begin{center}
\includegraphics[width=1.0\linewidth]{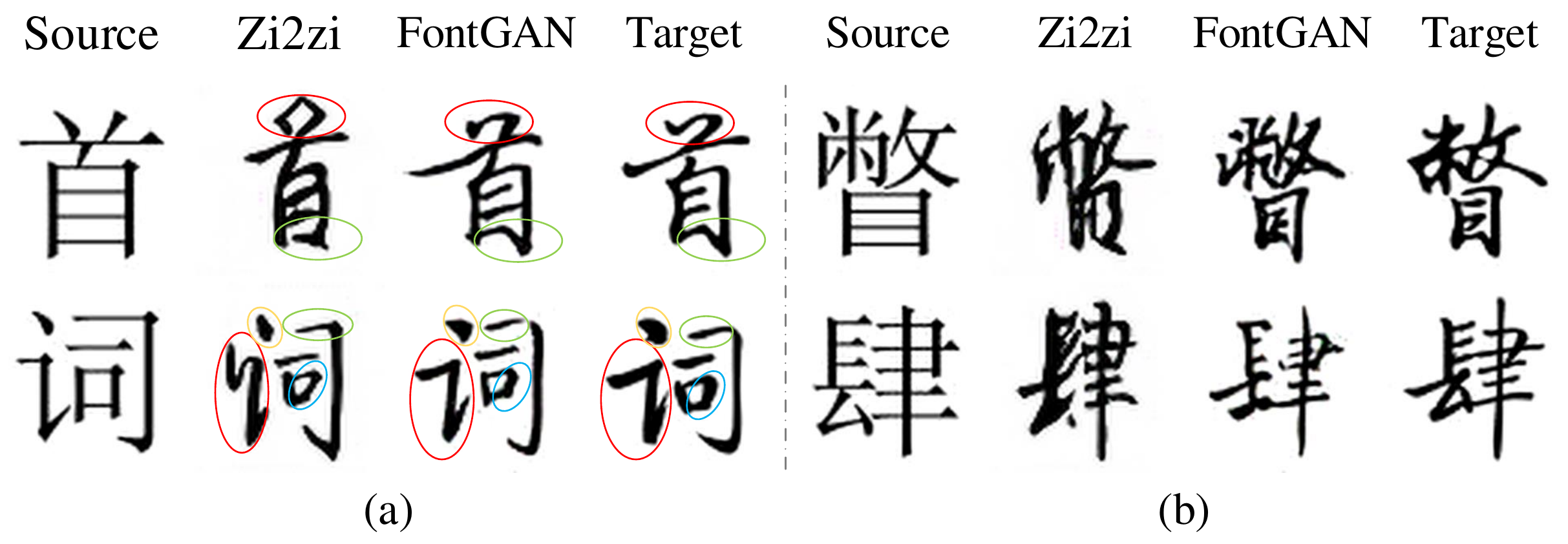}
\end{center}
   \caption{Comparisons of our method with zi2zi. (a) FontGAN does a better work of style consistency. (b) FontGAN can alleviate the problem of complex character generation.}
\label{fig:8}
\end{figure}

Fig.~\ref{fig:6} shows the stylization results for some challenging fonts such as calligraphy and handwriten character. Since the topology of these characters differ greatly from standard font, the difficulty of font translation is greatly increased. Moreover, a large number of joined-up characters and non-standard writings lead to the model that needs to balance style retention and content consistency. Experimental results show that our method can effectively alleviate this challenge. For example, as shown in the first column of Fig.~\ref{fig:6}, Rewrite and pix2pix do work poorly, while CGAN, zi2zi and our method can generate accurate results, but our results are closer to the target character. We further details the generated images in Fig.~\ref{fig:8} (a), these marked strokes indicate that our model can produce better results. Fig.~\ref{fig:8} (b) demonstrates that our method can alleviate the challenging of generating complex characters.

\begin{figure}
\begin{center}
\includegraphics[width=1.0\linewidth]{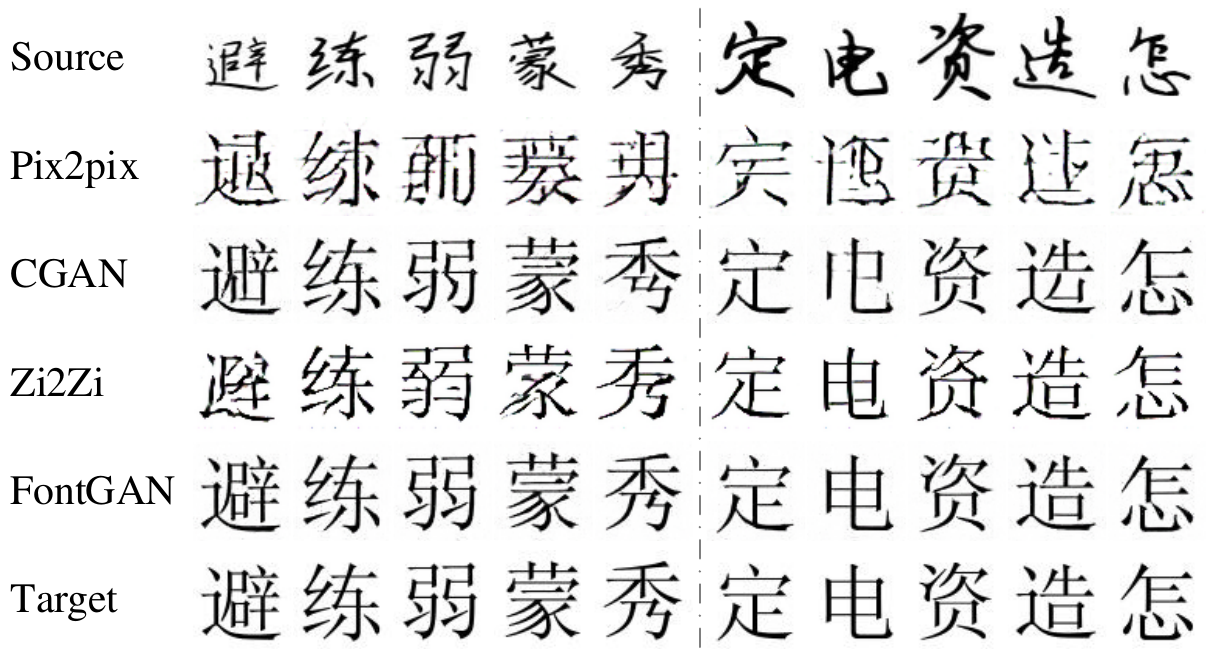}
\end{center}
   \caption{Character de-stylization comparisons. Two font styles are displayed, each consisting of five characters.}
\label{fig:7}
\end{figure}

\subsubsection{De-stylization}

With regard to character de-stylization, since rewrite does a very poorly work, we decide to remove this method in the next experiments. As shown in Fig.~\ref{fig:7}, pix2pix yields results that have SimSun style but the contents of the characters are completely confusing. The results obtained by CGAN and zi2zi can be slightly accepted, but there are still some problems of stroke missing or offset. Our method can accurately generate the desired characters.

In a word, our approach outperforms the state-of-the-art method in both character stylization and de-stylization, especially in more complicated situations, such as the fonts with large topology variations. This is because our model is able to accurately decouple the input image into content variables and style variables and further constrain these two types of variables through CPM and FCM.

\subsubsection{Transfer of Arbitrary Fonts}

\begin{figure}
\begin{center}
\includegraphics[width=1.0\linewidth]{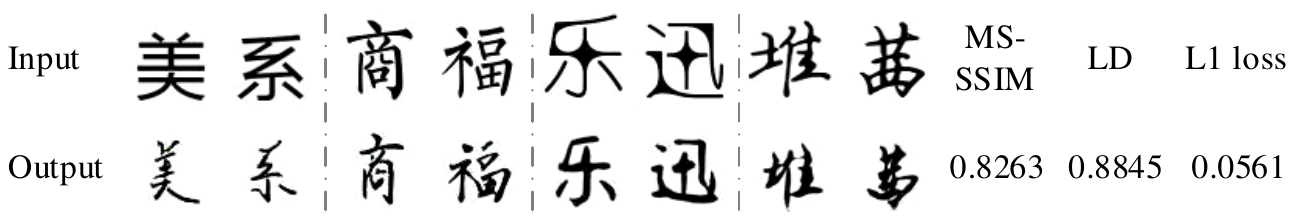}
\end{center}
   \caption{Many-to-many character translation. The results of four paired fonts are shown.}
\label{fig:15}
\end{figure}

We have achieved many-to-many character translation with one framework, which benefit from two reasons: (1) FontGAN decomposes the character into style representation and content representation, respectively; (2) It maps the content of all fonts to the same space. Specifically, we first obtain the content variable using $E_{c}$, and then feed it along with the style variable to the generator to synthesize the character of the specified style. Fig.~\ref{fig:15} shows that in addition to mapping standard font to other target fonts, our approach also enables mutual translation between target fonts.

\subsection{Quantitative Evaluation}

We select 10 fonts and each font contains 100 characters as test dataset. Table~\ref{table1} shows the stylization results, compared with zi2zi, our model achieves $2.62\%$ improvement in terms of MS-SSIM, $7.22\%$ improvement of LD and $10.54\%$ improvement of L1 loss, which indicate that our method outperforms the baseline in structural similarity, local smoothness and pixel-level similarity. As shown in Table~\ref{table2}, our method also achieves the best performance which outperforms zi2zi by a large margin ($6.81\%$, $41.94\%$ and $14.54\%$ improvements of MS-SSIM, LD and L1 loss, respectively). OCR accuracy also demonstrate the superiority of our model, especially in terms of content consistency.

\begin{table}
\caption{Comparison of different methods of stylization.}\smallskip
\centering
\resizebox{\columnwidth}{!}{
\smallskip\begin{tabular}{c|ccccc}
\hline
Methods & Rewrite & Pix2pix & CGAN & Zi2zi & FontGAN \\
\hline
MS-SSIM & 0.5084 & 0.8191 & 0.8114 & 0.8334 & \textbf{0.8552} \\
\hline
LD      & 5.2049 & 0.9703 & 0.8847 & 0.7925 & \textbf{0.7353} \\
\hline
L1 loss & 0.0908 & 0.0606 & 0.0555 & 0.0607 & \textbf{0.0543} \\
\hline
\end{tabular}
}
\label{table1}
\end{table}

\begin{table}
\caption{Comparison of different methods of de-stylization.}\smallskip
\centering
\resizebox{\columnwidth}{!}{
\smallskip\begin{tabular}{c|ccccc}
\hline
Methods & Rewrite & Pix2pix & CGAN & Zi2zi & FontGAN \\
\hline
MS-SSIM & - & 0.7802 & 0.8476 & 0.8225 & \textbf{0.8785} \\
\hline
LD      & - & 0.6908 & 0.4100 & 0.6013 & \textbf{0.3491} \\
\hline
L1 loss & - & 0.0644 & 0.0599 & 0.0571 & \textbf{0.0488} \\
\hline
OCR     & - & 0.6035 & 0.7456 & 0.7440 & \textbf{0.8282} \\
\hline
\end{tabular}
}
\label{table2}
\end{table}

\subsection{Inference with New Fonts}

\begin{figure}
\begin{center}
\includegraphics[width=1.0\linewidth]{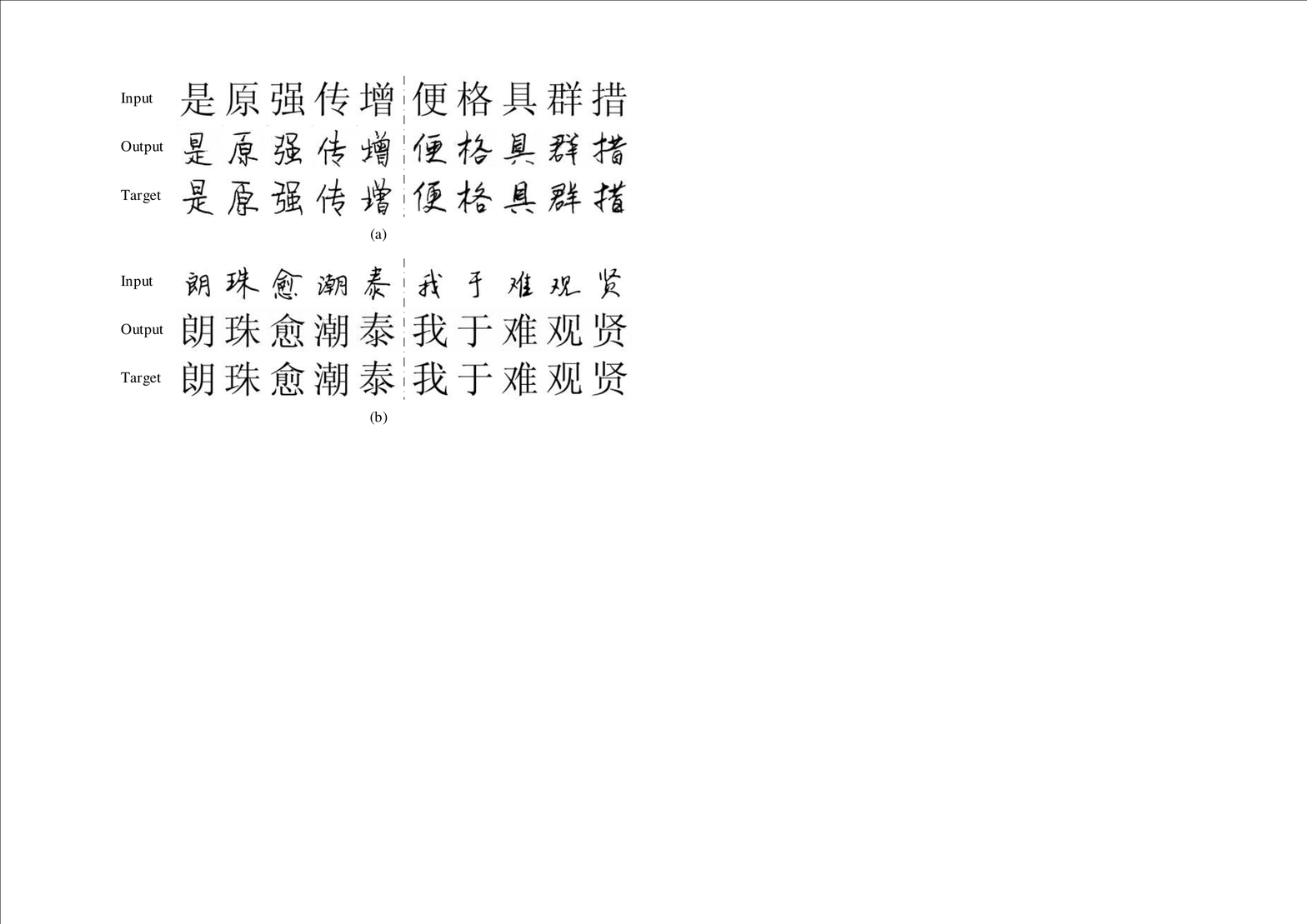}
\end{center}
   \caption{Inference with new fonts. (a) It shows the stylization results with new fonts. (b) It show the de-stylization results with new fonts.}
\label{fig:13}
\end{figure}

We have further considered extending our model to novel fonts. For character stylization, it takes only a few minutes to fine-tune the trained model with limited data to achieve satisfactory results (see Fig.~\ref{fig:13} (a)). With regard to character de-stylization, our model can directly map the input characters into SimSun font without training (see Fig.~\ref{fig:13} (b)).

\subsection{Ablation Experiments}

We also perform a series of ablation experiments to validate the effectiveness of FCM and CPM. Moreover, style/content verification experiments are also introduced.

\subsubsection{Effectiveness of FCM.}

\begin{figure}
\begin{center}
\includegraphics[width=1.0\linewidth]{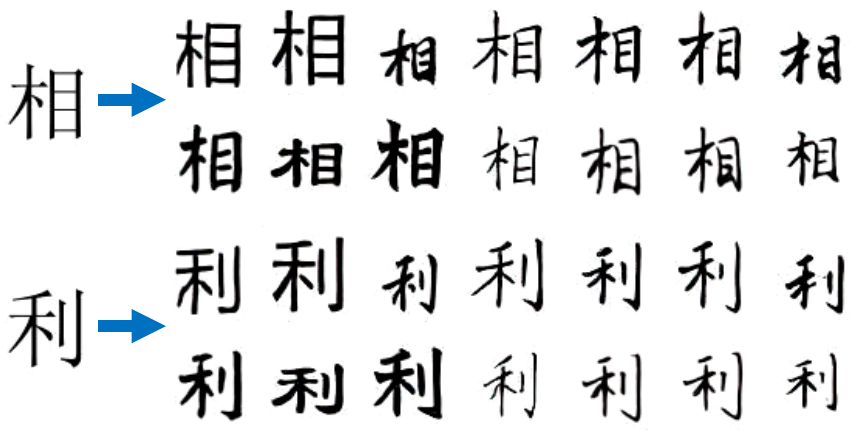}
\end{center}
   \caption{Examples obtained by sampling. Fourteen styles of each character are displayed.}
\label{fig:10}
\end{figure}

FCM can further improve the quality of the generated images, but the improvement is not obvious. As shown in Fig.~\ref{fig:9}(a), the model trained with FCM is more advantageous in terms of style maintenance of character details. Moreover, we introduce FCM to obtain the specified style variables by sampling the Gaussian distribution. The final result is then generated by combining style variables and content variables (as shown in Fig.~\ref{fig:10}).

\subsubsection{Effectiveness of CPM.}

\begin{figure}
\begin{center}
\includegraphics[width=1.0\linewidth]{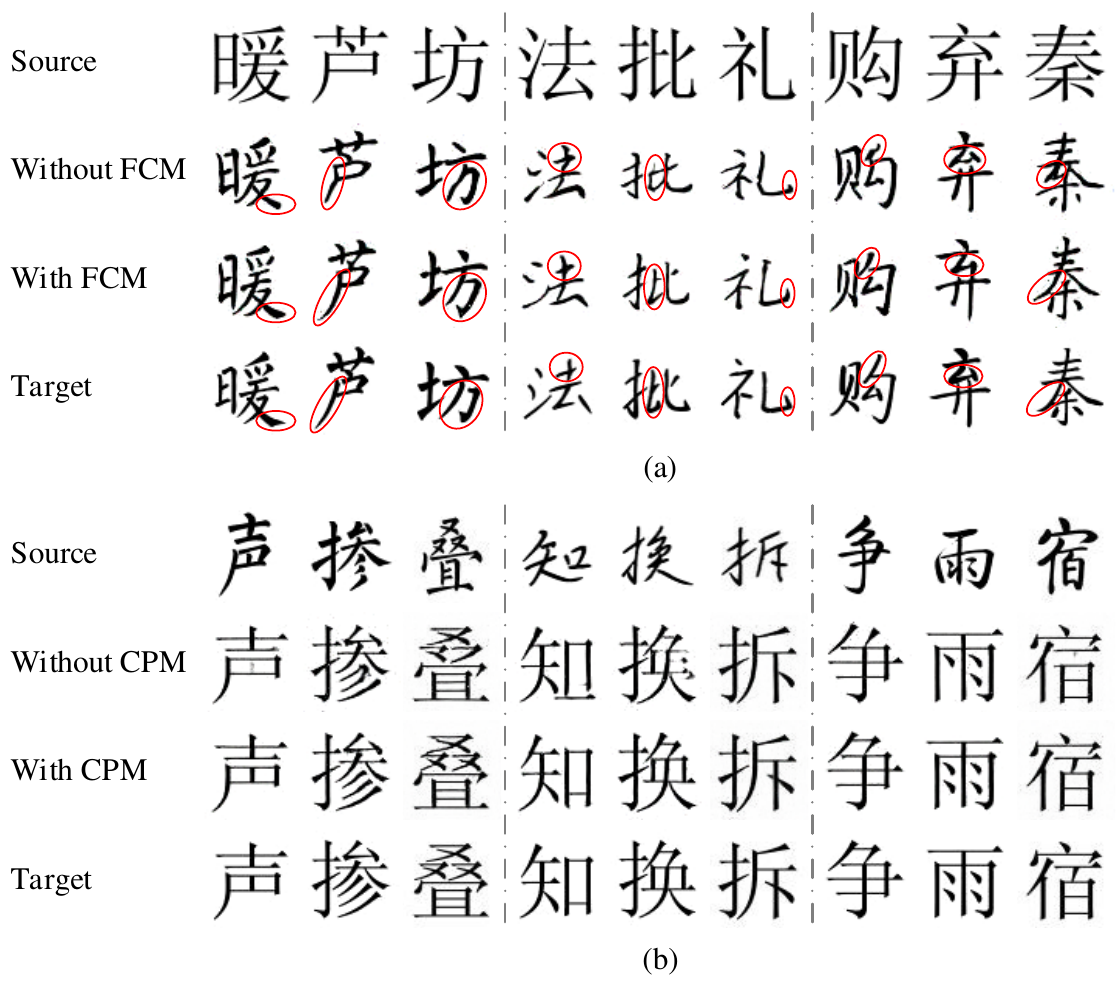}
\end{center}
   \caption{Ablation experiments. (a) Analysis of FCM. FCM enables our model to do a better work in style maintenance, especially in terms of detail. (b) Analysis of CPM. CPM can alleviate the problem of stroke deficiency in character de-stylization. It greatly improves the quality and accuracy of the generated results.}
\label{fig:9}
\end{figure}

CPM is mainly used to alleviate the problem of stroke deficiency or confusion during character de-stylization. As illustrated in Fig.~\ref{fig:9}(b), we first train our model without CPM, and then evaluate it on test dataset. Our model can generally yields satisfactory results, but in some cases the strokes are incomplete. On the contrary, by introducing CPM, the generation quality is significantly improved.

\subsection{Validation Experiments}

It is worth noting that we get some wrong characters or the characters that never existed by manual operation (e.g. the first column of Fig.~\ref{fig:16} (a) or the first row of Fig.~\ref{fig:16} (b)). Experimenting on these characters can strongly prove the effectiveness of style variables and content encoder.

\subsubsection{Effectiveness of Style Variables}

We select 2 characters to provide content variables, and then apply 10 style variables to these 2 characters respectively. The final generated results are shown in Fig.~\ref{fig:16} (a). We can find that the style variables is able to work on new and different content and yield compelling results, which demonstrates that style variables are not affected by the content of the characters.

\subsubsection{Effectiveness of Content Encoder}

\begin{figure}
\begin{center}
\includegraphics[width=1.0\linewidth]{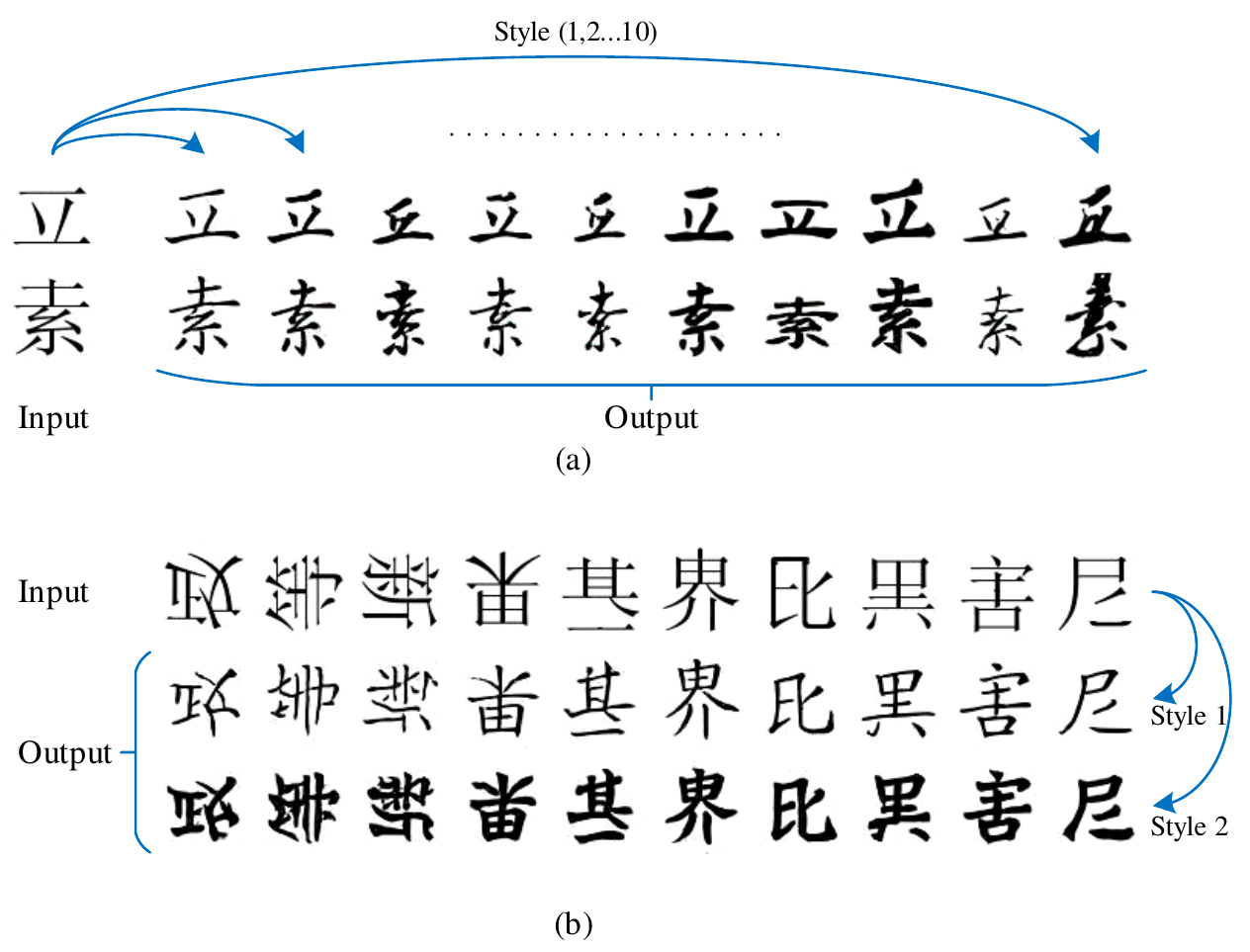}
\end{center}
   \caption{Validation experiments. (a) Effectiveness of style variables. Two characters are used to provide the reference content variables, which combine with ten style variables to generate characters. The illustrated results show that style variables are not affected by the content of the characters. (b) Effectiveness of content encoder. Ten character contents are used to generate the characters with the specified style, which prove that the content encoder can effectively encode content information.}
\label{fig:16}
\end{figure}

We adopt 2 fonts as the reference style variables. The wrong characters are first fed into content encoder to get the content variables, which are then combined with the reference style variables to generate the final images. As shown in Fig.~\ref{fig:16}(b), the generated contents are highly consistent with the original contents. The content encoder is able to accurately and completely map the contents of the original character to the latent space, which is not influenced by the stroke layout of the character.

\subsection{Limitations and Discussion}

\begin{figure}
\begin{center}
\includegraphics[width=1.0\linewidth]{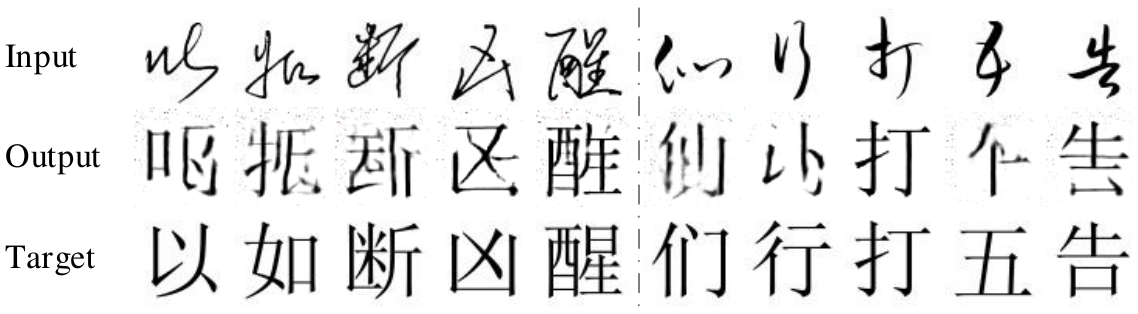}
\end{center}
   \caption{Some failures of our method. Character de-stylization of inference with new complicated fonts.}
\label{fig:14}
\end{figure}

Although our method is able to yield desirable results in most cases, there are also some failures. For example, when conducting de-stylization inference with new fonts, if the style of the new font is very complicated, the generated result is often unsatisfactory (Fig.~\ref{fig:14}). We think this is because these characters and standard characters vary greatly in topology, which affects the performance of the content encoder.

\section{Conclusion}

In this paper, we propose a unified framework for Chinese character stylization and de-stylization. We formulate the de-stylization process as a many-to-one image translation task. Unlike typical character recognition, our method start with the perspective of image translation, which can directly map various fonts into a standard font (e.g. SimSun). It will facilitate text digitization and subsequent editing and sharing. In addition, we decouple the character into style representation and content representation, which is more conducive to learning the corresponding feature representation in the deep space. Furthermore, font consistency module (FCM) and content prior module (CPM) are proposed. FCM helps our model to learn font style more accurately, and can obtain style representation through Gaussian sampling without reference image. CPM improves the stability and accuracy of the character de-stylization and alleviates the problem of stroke deficiency. Both qualitative and quantitative results demonstrate the effectiveness of our approach.

{\small
\bibliographystyle{ieee_fullname}
\bibliography{mine}
}

\end{document}